\documentclass[10pt,twocolumn,letterpaper]{article}

\usepackage{icb}
\usepackage{times}
\usepackage{epsfig}
\usepackage{graphicx}
\usepackage{amsmath}
\usepackage{amssymb}

\usepackage[hyphens]{url}
\usepackage[flushleft]{threeparttable}
\usepackage{tablefootnote}
\usepackage{booktabs}
\usepackage[export]{adjustbox}
\usepackage{array}
\newcolumntype{P}[1]{>{\centering\arraybackslash}p{#1}}
\newcolumntype{M}[1]{>{\centering\arraybackslash}m{#1}}
\usepackage{dblfloatfix}
\usepackage{algorithmic}
\usepackage{algorithm}
\usepackage{varwidth}
\usepackage[caption=false,font=footnotesize]{subfig}

\icbfinalcopy 

\ificbfinal\fi
\begin{document}

\title{Fingerprint Match in Box}

\author{Joshua J. Engelsma\\
Michigan State University\\
East Lansing, MI, USA\\
{\tt\small engelsm7@msu.edu}
\and
Kai Cao\\
Michigan State University\\
East Lansing, MI, USA\\
{\tt\small kaicao@cse.msu.edu}
\and
Anil K. Jain\\
Michigan State University\\
East Lansing, MI, USA\\
{\tt\small jain@cse.msu.edu}
}

\maketitle
\thispagestyle{empty}

\begin{abstract}
We open source fingerprint Match in Box, a complete end-to-end fingerprint recognition system embedded within a 4 inch cube. Match in Box stands in contrast to a typical bulky and expensive proprietary fingerprint recognition system which requires sending a fingerprint image to an external host for processing and subsequent spoof detection and matching. In particular, Match in Box is a first of a kind, portable, low-cost, and easy-to-assemble fingerprint reader with an enrollment database embedded within the reader's memory and open source fingerprint spoof detector, feature extractor, and matcher all running on the reader's internal vision processing unit (VPU). An onboard touch screen and rechargeable battery pack make this device extremely portable and ideal for applying both fingerprint authentication (1:1 comparison) and fingerprint identification (1:N search) to applications (vaccination tracking, food and benefit distribution programs, human trafficking prevention) in rural communities, especially in developing countries. We also show that Match in Box is suited for capturing neonate fingerprints due to its high resolution (1900 ppi) cameras. 

\end{abstract}

\section{Introduction}
Automated fingerprint identification systems (AFIS) are now prevalent around the globe providing an accurate and widely acceptable method for authentication (1:1 match) or identification and de-duplication (1:N search) of individuals~\cite{handbook}. To date, fingerprint recognition systems have been successfully deployed into a plethora of applications including healthcare access, financial transactions, forensics, border crossing security, national ID systems, and mobile device access and payments~\cite{handbook}. Despite widespread deployment of fingerprint recognition systems in place of ID cards and passwords, they still need to overcome some hurdles before their further deployments into several niche domains, especially pertaining to applications in developing countries. These limitations include spoofing vulnerabilities, insufficient resolution for neonate fingerprinting, lack of portability, difficult customization, and high cost of hardware and software. 

\newcommand{\specialcell}[2][c]{%
  \begin{tabular}[#1]{@{}c@{}}#2\end{tabular}}
  \newcommand{\tabitem}{~~\llap{\textbullet}~~}

\begin{figure}[t]
\begin{center}
\includegraphics[scale=0.45]{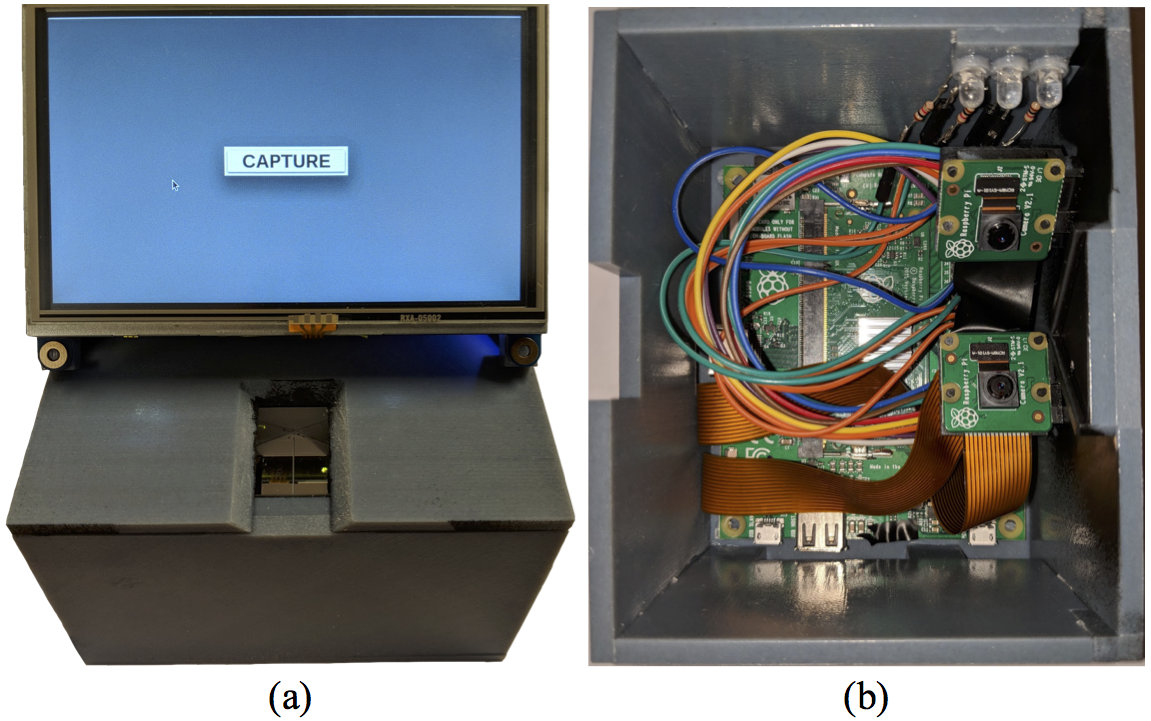}
\caption{Match in Box. (a) A working prototype of Match in Box. Match in Box is a complete, open-source, end-to-end fingerprint recognition system (with 1900 ppi fingerprint acquisition, spoof detector, feature extractor, template storage, and 1:N search) embedded within a portable (rechargeable battery and touch screen) 4-inch cube. (b) Assembly of Match in Box requires only low-cost (\$400 USD), ubiquitous, off-the-shelf components, enabling its easy build and customization for niche applications, particularly in inaccessible locations in developing countries.
}
\label{fig:intro_fig}
\end{center}
\end{figure} 

\begin{figure*}[t]
\begin{center}
\includegraphics[scale=0.58]{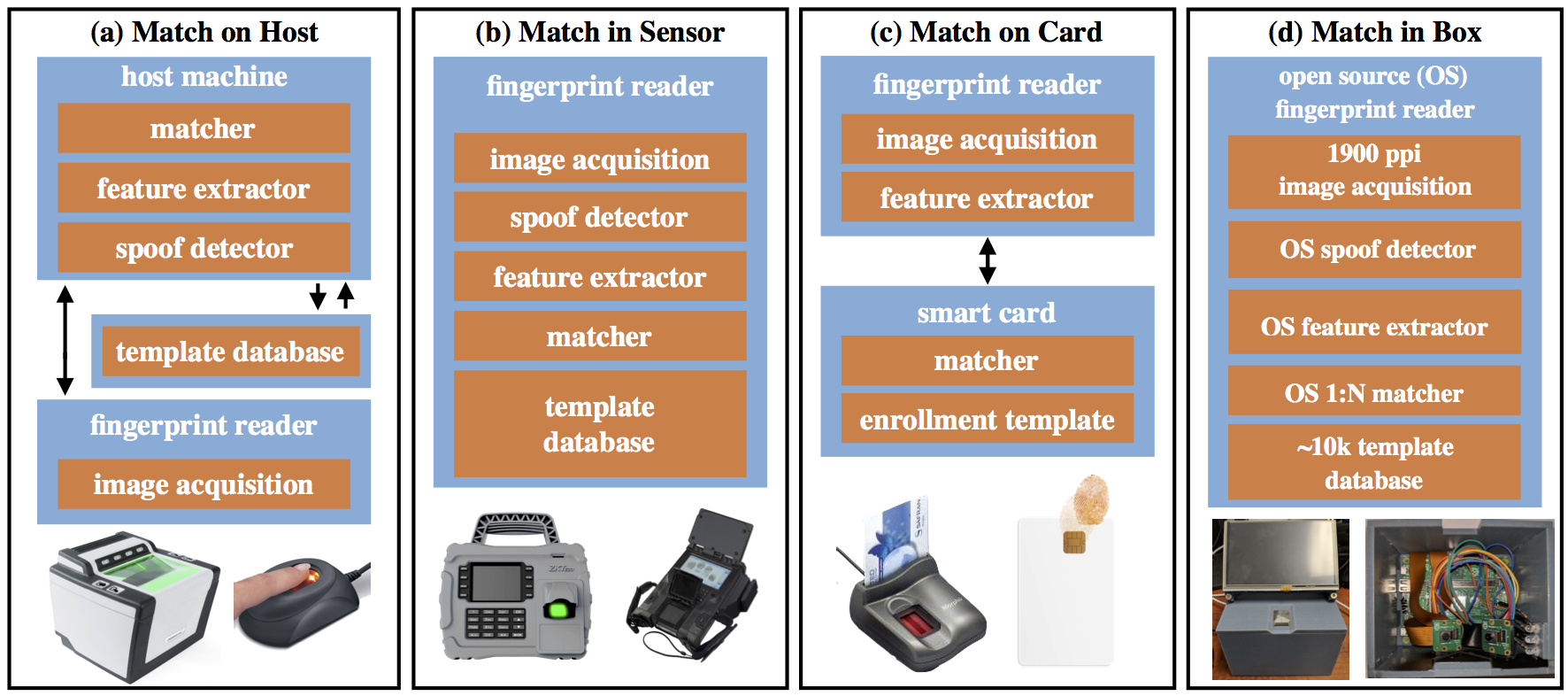}
\caption{Schematic diagrams and example commercial fingerprint recognition system architectures. (a) Match on Host, (b) Match in Sensor, (c) Match on Card, and (d) the proposed Match in Box, the first open source implementation following the Match in Sensor architecture, allowing customization to fit application specific needs, particularly in developing countries. Figures for commercial products retrieved from~\cite{prior1, prior2, prior3, prior4, prior5, prior6}.
}
\label{fig:prior_work}
\end{center}
\vspace{-1.5em}
\end{figure*} 

\subsection{Spoofing Vulnerability}
Fingerprint recognition systems are known to be susceptible to circumvention through ``{\it spoofing}" attacks\footnote{Fingerprint spoof attacks are a subset of ``{\it presentation attacks}" defined in ISO standard IEC 30107-1:2016(E) as ``{\it presentation to the biometric data capture subsystem with the
goal of interfering with the operation of the biometric system.}"}. Fingerprint spoofing is a nefarious act in which a hacker mimics the identity of an unsuspecting victim (through replicating the victim's fingerprint with materials such as silicone, gelatin, and wood glue) in order to gain access to the victim's biometric protected data~\cite{gummy_fingers, burned}. Any failure of a fingerprint recognition system to automatically detect and flag spoof attacks prior to authentication could result in loss of personal data and benefits, or preventing proper identification of a person of interest (by obfuscating a person's true identity).

Given the requirements of accurately detecting fingerprint spoof attacks in most applications, a number of techniques involving both hardware and software have been proposed~\cite{spoofing_survey}. In hardware based approaches, additional sensors embedded within the fingerprint readers output/provide features such as skin color, sub-dermal fingerprints, blood flow, odor, and heartbeat which can be used to distinguish live fingers from spoof fingers~\cite{spoofing_survey, raspireader, com1, com2, com3, odor, blood_flow, rowe1, oct}. In contrast to hardware based solutions, software based techniques do not require special sensors or special illumination within the fingerprint reader. Instead, software based approaches operate by extracting textural~\cite{texture0}, anatomical~\cite{pores}, physiological~\cite{perspiration1}, or learned~\cite{tarang, CNN2} features from images being output by the reader for the purpose of fingerprint recognition (i.e. the same image is used for both spoof detection and subsequent recognition). To date, much work remains to be done in spoof detection research, especially in detecting spoofs not used during training of the spoof detectors~\cite{inter1, inter2}.

\subsection{Insufficient Fingerprint Resolution}

A limitation of commercial fingerprint readers is that most only produce 500 ppi fingerprint images, because of the cost and the targeted adult end users. However, as noted by Jain {\it et al.} in~\cite{infantfingerprint}, higher fingerprint resolution ($> 1000$ ppi) is necessary to capture the minute fingerprint details of a neonate or an infant. Neonate biometric recognition, especially involving fingerprints, is an area of increasing interest by governments (e.g. Aadhaar program~\cite{aadhaar0, aadhaar}), international agencies (e.g. various UN agencies such as WFP~\cite{wfp}, and UNCHR~\cite{unhcr}) and NGOs (e.g. Bill and Melinda Gates Foundation~\cite{bill}). While commercial 1000 ppi fingerprint reader solutions do exist~\cite{NEC, crossmatch}, they can cost upwards of over \$1000 USD, negating the practicality of their utility in capturing neonate fingerprints, especially in the developing world.

\begin{table*}[t]
 \centering
\begin{threeparttable}
\begin{tabular}{ |c||c|c|c|c|c|c|c|}
 \multicolumn{8}{c}{Table 1: Match in Box vs RaspiReader} \\
 \hline
 Reader & \specialcell{Capture \\ Time \\(seconds)} & \specialcell{Capture \\ Area \\(pixels)} & \specialcell{Native $x$ \\ Resolution \\(ppi)} & \specialcell{Native $y$\\ Resolution \\(ppi)} & \specialcell{Spoof \\Detector} & \specialcell{Feature Extractor \\ \& Matcher} & \specialcell{Portability} \\
 \hline
 \hline
 RaspiReader~\cite{raspireader} & 1.5 & 290x267 & 1594 - 2480 & 2463 - 3320 & on host & on host & low\\
 \hline
 Match in Box & \textbf{0.75} & \textbf{285x357} & \textbf{1917-2962} & \textbf{2900-3768} & \textbf{embedded} & \textbf{embedded} & \textbf{high}\\
  \hline
\end{tabular}
\end{threeparttable}
\end{table*}

\subsection{Portability}

Many fingerprint recognition systems have limited portability due to their reliance on multiple physical hardware modules. In a typical ``Match in Host" (Fig. \ref{fig:prior_work} (a))~\cite{synaptics} system, a fingerprint reader acquires the fingerprint image which is then transferred to a host to perform spoof detection, feature extraction, template storage, and matching. Although the host offers choice of processors and memory, the Match in Host design suffers from two main drawbacks: (i) sensitive biometric data needs to be transferred from the fingerprint reader to the host and (ii) the system is costly, bulky, and immobile. These limitations are especially prohibitive to field operations in developing countries (such as vaccination and health tracking of children, banking, and benefits distributions) where low cost and portability are essential given the lack of financial resources, electric power and network reliability. 

To solve the limitations inherent to Match in Host architectures, two newer generations of fingerprint recognition architectures emerged. The first of these architectures has been termed ``Match in Sensor" (Fig. \ref{fig:prior_work} (b))~\cite{synaptics} or alternatively described as operating in ``Embedded Mode"~\cite{prior3, prior4,lumidigm1}. Match in Sensor operates entirely (fingerprint acquisition, feature extraction, template storage, matching, spoof detection) within the processor and memory, embedded within the fingerprint reader. The advantages of this newer paradigm are; (i) higher security, since fingerprint data does not need to be transmitted and (ii) greater portability. 

Another fingerprint recognition architecture, known as ``Match on Card" (Fig. \ref{fig:prior_work} (c))~\cite{matchoncard, gemalto}, has now been gaining traction. There are two instances of the Match on Card paradigm (dependent on whether the fingerprint reader is on the card or not). If the fingerprint reader is {\it not} on the card, a user simply inserts a card into a specialized fingerprint reader which (i) acquires the user's fingerprint, (ii) extracts a template from the fingerprint image, and (iii) securely transmits the template to the card where the template can be securely stored as an enrollment template or used for matching (authentication) directly on the smart card. If the fingerprint reader is already on the card, then fingerprint image acquisition, feature extraction and matching can all be performed directly on the card.

Both the Match in Sensor and the Match on Card paradigms solve {\it some} of the limitations inherent to Match in Host. However, the commercial products available for Match in Sensor and Match on Card still have several limitations of their own, especially as it pertains to applications in developing countries involving identification of tens of thousands of individuals. These limitations include: inability to customize the black box systems, high-cost, insufficient resolution for neonate fingerprints, and spoofing vulnerabilities.


\begin{table*}[t]
\begin{center}
\begin{threeparttable}
\label{tab:parts}
\resizebox{\textwidth}{!}{%
\begin{tabular}{ |c||M{5.5cm}|c|c|}
 \multicolumn{4}{c}{{\tiny Table 2: Primary Components Used to Construct Match in Box. Total cost is \$400 USD (as of April 1, 2018)}} \\
 \hline
{\tiny Component Image} & {\tiny Name and Description} & {\tiny Quantity} & {\tiny Cost (USD)}\\
\hline
 \includegraphics[valign=m,scale=0.5]{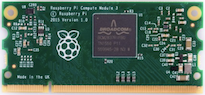} & {\tiny \textbf{Raspberry Pi Compute Module 3 Lite~\cite{cm3}:}  A single board computer (SBC) with 1.2 GHz 64-bit quad-core CPU, 1 GB RAM, MicroSDHC storage, and Broadcom VideoCore IV Graphic card} & {\tiny1} & {\tiny \$29.95}\\
 \hline
 \includegraphics[valign=m,scale=.2]{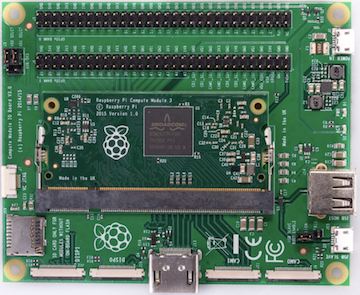}  & {\tiny \textbf{Raspberry Pi Compute I/O Board~\cite{ioboard}:} Compute Module breakout board} & {\tiny 1} & {\tiny \$114.95}\\
  \hline
 \includegraphics[valign=m,scale=0.05]{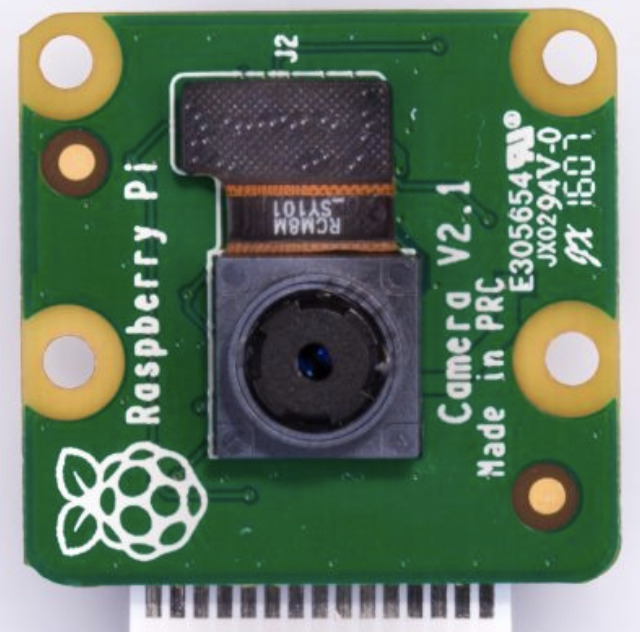} & {\tiny \textbf{Pi Camera v2~\cite{picamera}:} 8 Megapixel, 30 frames per second} & {\tiny 2} & {\tiny \$29.95}\\
  \hline
 \includegraphics[valign=m,scale=0.075]{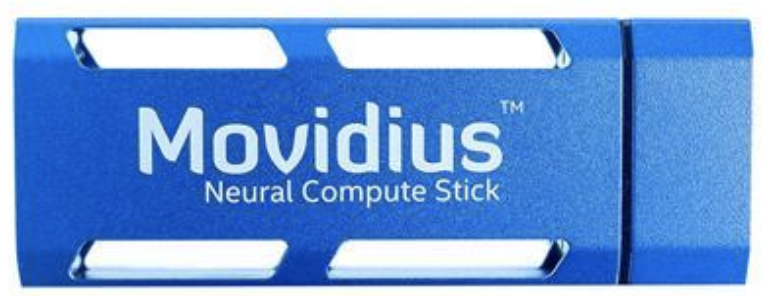}  & {\tiny \textbf{Intel Neural Compute Stick~\cite{neural}:} Intel Movidius VPU processor} & {\tiny 1} & {\tiny \$80.43}\\
  \hline
\includegraphics[valign=m,scale=0.05]{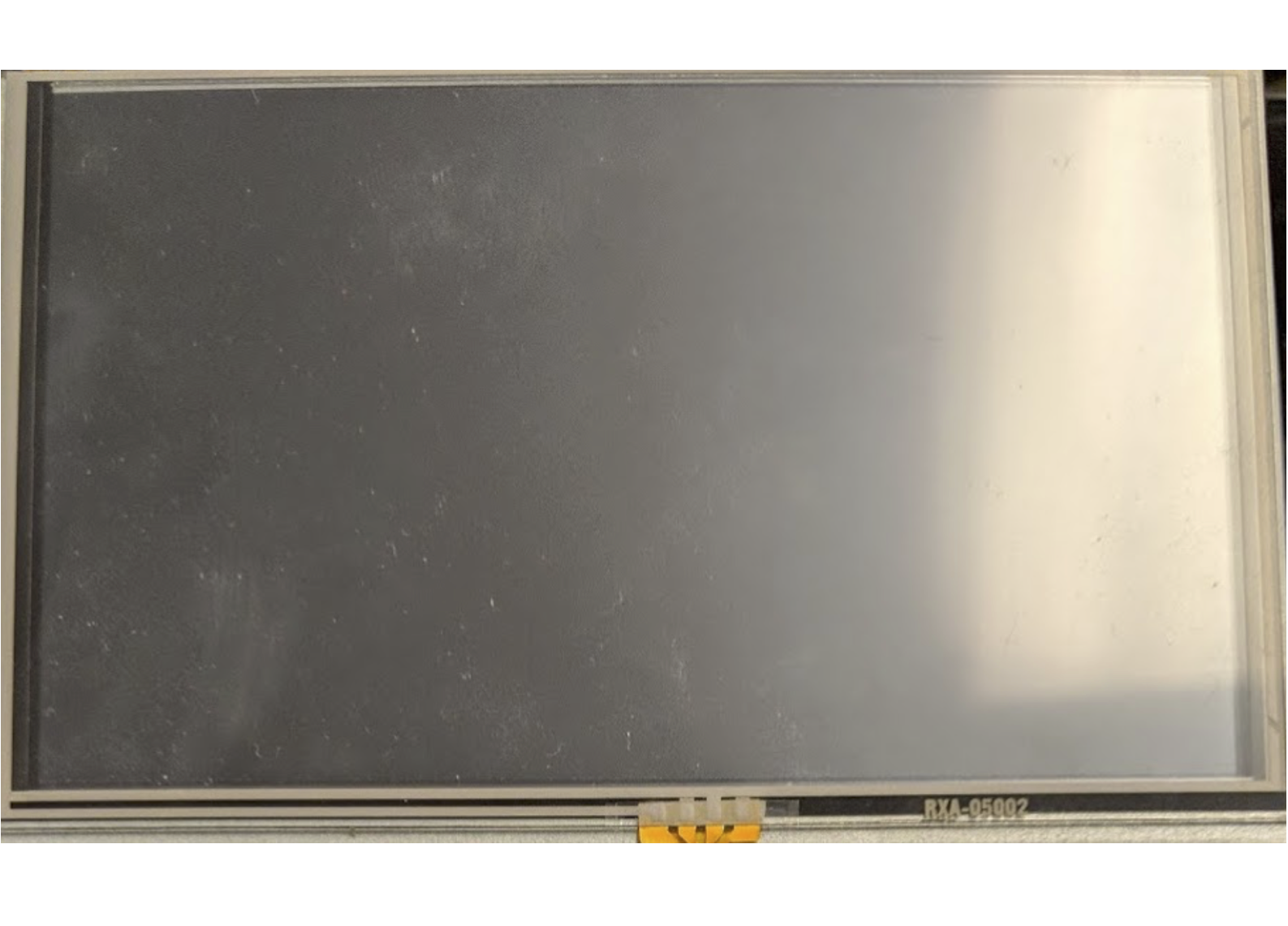} & {\tiny \textbf{Touch Screen~\cite{tft}:} 5-inch TFT touch screen} & {\tiny1} & {\tiny \$74.95}\\
  \hline
\includegraphics[valign=m,scale=0.5]{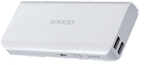}  & {\tiny \textbf{Rechargeable Battery~\cite{battery}:}~ 10,000 mAh lithium ion battery} & {\tiny 1} & {\tiny \$39.95}\\
\hline
 
\end{tabular}}
\end{threeparttable}
\end{center}
\vspace{-1.0em}
\end{table*} 

In summary, the advances in spoof detection research, fingerprint readers, and system mobility have come a long ways, however, gaps still exist, especially with respect to utilizing fingerprint recognition technologies in many emerging and critical applications in the developing world, including: delivering nutrition to undernourished children~\cite{childhunger}, improving child vaccination coverage~\cite{immunizationcoverage}, and government distribution programs~\cite{aadhaar, grain}. Given these limitations of available commercial fingerprint recognition systems in many applications specifically relevant to developing countries, we open source fingerprint Match in Box, a cost-effective, customizable, mobile, spoof resistant, and 1900 ppi end-to-end fingerprint recognition system aimed at filling these voids. In particular, the proposed Match in Box makes several major customizations and improvements to RaspiReader~\cite{raspireader} (Table 1), including:

 \begin{itemize}
\item Reducing fingerprint image acquisition time and increasing the native fingerprint resolution.

\item Embedding a state-of-the-art fingerprint spoof detector, feature extractor, and matcher directly onto the Match in Box processor through the addition of a Vision Processing Unit (VPU). Enrollment templates of up to 10,000 subjects can be stored in the Match in Box internal memory storage.

\item Complete mobility and portability through the addition of a touch screen, intuitive GUI, and rechargeable battery pack. The touch screen allows for easy enrollment, authentication and metadata access of enrolled subjects.

\item Experimental results demonstrating capability of Match in Box for (i) spoof detection, (ii) fingerprint matching, and (iii) potential for state-of-the-art neonate fingerprint recognition.
  
\end{itemize}

\section{Match in Box Hardware}

In assembling Match in Box, the following hardware design choices (including image acquisition hardware, processor and memory requirements for Match in Sensor capabilities, and peripherals for portability) are made to best meet our objective of prototyping an open source, spoof resistant, high-resolution, portable, customizable, low-cost, end-to-end fingerprint recognition system. Table 2 lists all the hardware components that were used to assemble Match in Box. Except for the casing that was custom fabricated using a Stratasys Objet350 Connex~\cite{connex} 3D printer, all other components to construct the reader are available off-the-shelf.

\subsection{Image Acquisition Hardware}
\begin{figure}[b]
\begin{center}
\includegraphics[scale=0.5]{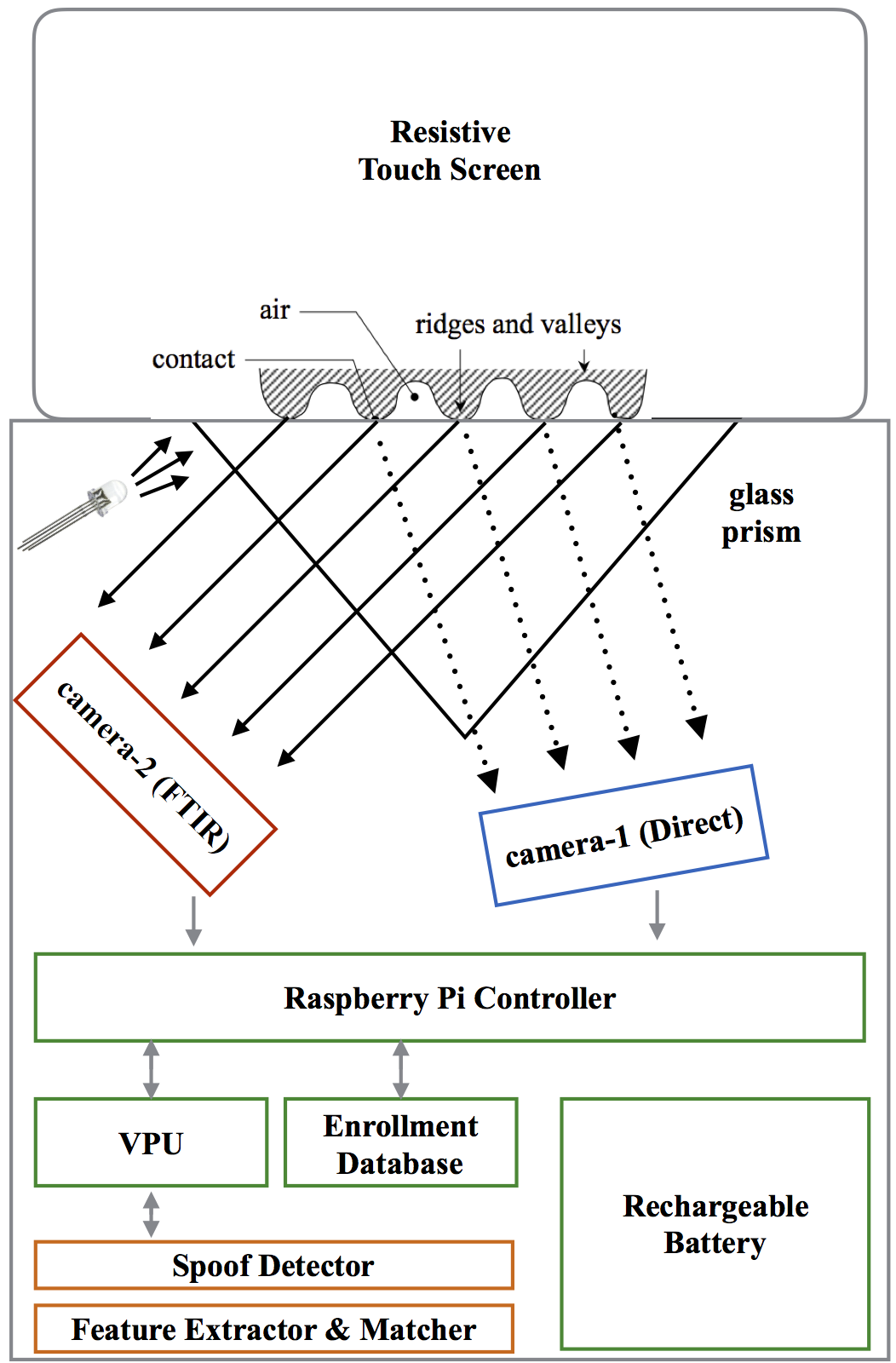}
\caption{Schematic diagram of Match in Box.
}
\label{fig:schematic}
\end{center}
\vspace{-1.5em}
\end{figure}

To make Match in Box spoof resistant, we follow the success of RaspiReader in using two cameras for fingerprint image acquisition. In particular, one camera captures a colorful, {\it direct image} of the finger in contact with the glass platen (useful for spoof detection) while the second camera captures a high contrast, {\it frustrated total internal reflection image} (useful for both spoof detection and subsequent fingerprint matching) (Figs. \ref{fig:schematic}, \ref{fig:images}). Unlike RaspiReader, Match in Box uses the Raspberry Pi Compute Module 3 Lite (CM3L) to capture two still-images or video streams simultaneously. 

The CM3L Raspberry Pi is a single board computer (SBC) which, when used in conjunction with the Raspberry Pi Compute Module I/O board and two 8 MP Raspberry Pi v2.0 cameras, is capable of capturing two still-images or video streams simultaneously in 0.75 seconds. Furthermore, the 8 MP cameras increase the fingerprint pixel resolution of Match in Box to 1917 ppi - 2962 ppi in the x-axis and 2900 ppi - 3768 ppi in the y-axis. This extremely high native resolution makes this affordable fingerprint recognition system extremely well suited for neonate fingerprint recognition.

\subsection{Processors and Memory}

An extremely vital and useful feature of Match in Box is its capability to perform fingerprint spoof detection and authentication (1:1 comparison) or identification (1:N search) directly on-board the fingerprint reader. To enable this functionality, Match in Box is equipped with a 32 Gigabyte SD card for storing approximately 10,000 fingerprint templates and associated user meta-data on device. Additionally, the processing speed of Match in Box is accelerated through the addition of the Intel Movidius Neural Compute Stick (NCS). The NCS is a low-power vision processing unit (VPU) designed with 12 specialized vector processing units, called SHAVES, for accelerating deep learning on the edge. The NCS is compatible with popular deep learning libraries Tensorflow and Caffe and is connected to USB. As such, the NCS is also compatible with the CM3L Raspberry Pi hardware and our Convolutional Neural Network (CNN) based spoof detection and fingerprint feature extraction software modules (both implemented in Tensorflow). While, one CNN inference of our ``light-weight" spoof detection model (MobileNet) takes 475 milliseconds to run on the Match in Box CPU, an inference on the Match in Box VPU (NCS) takes a mere 58 milliseconds, an almost ten-fold speedup.

\subsection{Portability Peripherals}

Finally, in order to make Match in Box a completely portable and standalone system, we add a rechargeable battery pack (with the capability of powering a Raspberry Pi for up to 15 hours) and a 5-inch, 800x480 thin film transistor (TFT) touch screen. 

All of these primary components are assembled together in accordance with the schematic diagram shown in Figure \ref{fig:schematic}. Given a basic knowledge of circuits and programming, the entire device can be fabricated in under one hour. Upon completion of assembly, Match in Box is ready for fingerprint image acquisition, spoof detection, feature extraction, and matching in any location.

\section{Match in Box Software}

The Match in Box software is comprised of four main modules: fingerprint image acquisition, fingerprint spoof detection, fingerprint feature extraction and matching (1:1 verification and 1:N search), and a Graphical User Interface (GUI) for enrolling users and subsequently displaying user meta-data. Each of these modules are further explained in the following subsections. 

\subsection{Fingerprint Acquisition}

\begin{figure}[h]
  \centering
  \subfloat[]{\includegraphics[scale=.375]{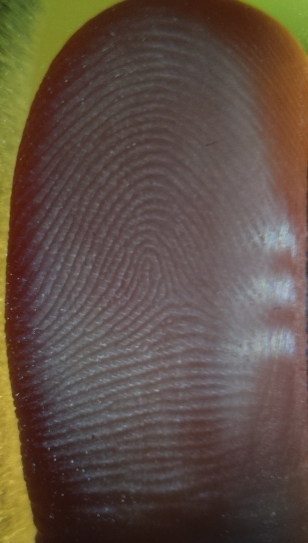}\label{fig:josh_direct}}
  \hfill
  \subfloat[]{\includegraphics[scale=.375]{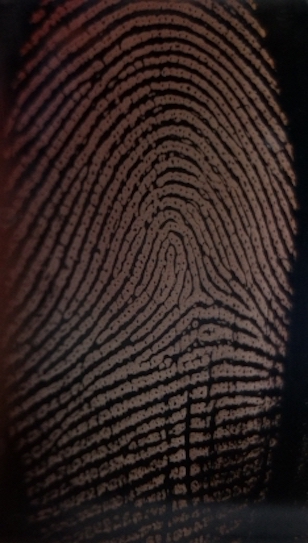}\label{fig:josh_ftir}}
  \caption{Match in Box captures two fingerprint images of a user's finger from two different cameras. (a) Direct image of the finger for spoof detection; (b) high-contrast FTIR image for both spoof detection and matching.}
  \label{fig:images}
 \vspace{-1.0em}
\end{figure}

The primary software involved in fingerprint image acquisition is the calibration of a raw FTIR image (e.g. Fig. \ref{fig:josh_ftir}) into a processed, grayscale FTIR image which can be used for fingerprint matching. Following the same steps as reported in~\cite{raspireader}, a raw FTIR fingerprint image is converted into a grayscale matching image by (i) RGB to grayscale conversion, (ii) contrast enhancement via histogram equalization, (iii) perspective transformation, and (iv) scaling to the required fingerprint resolution (500 ppi for adults or 1900 ppi for neonates). As in~\cite{raspireader}, the perspective transformation and scaling parameters are obtained using a 2D printed checkerboard pattern to estimate the source and destination coordinate pairs needed to estimate the transformation parameters.

\begin{figure*}[t]
\begin{center}
\includegraphics[scale=0.49]{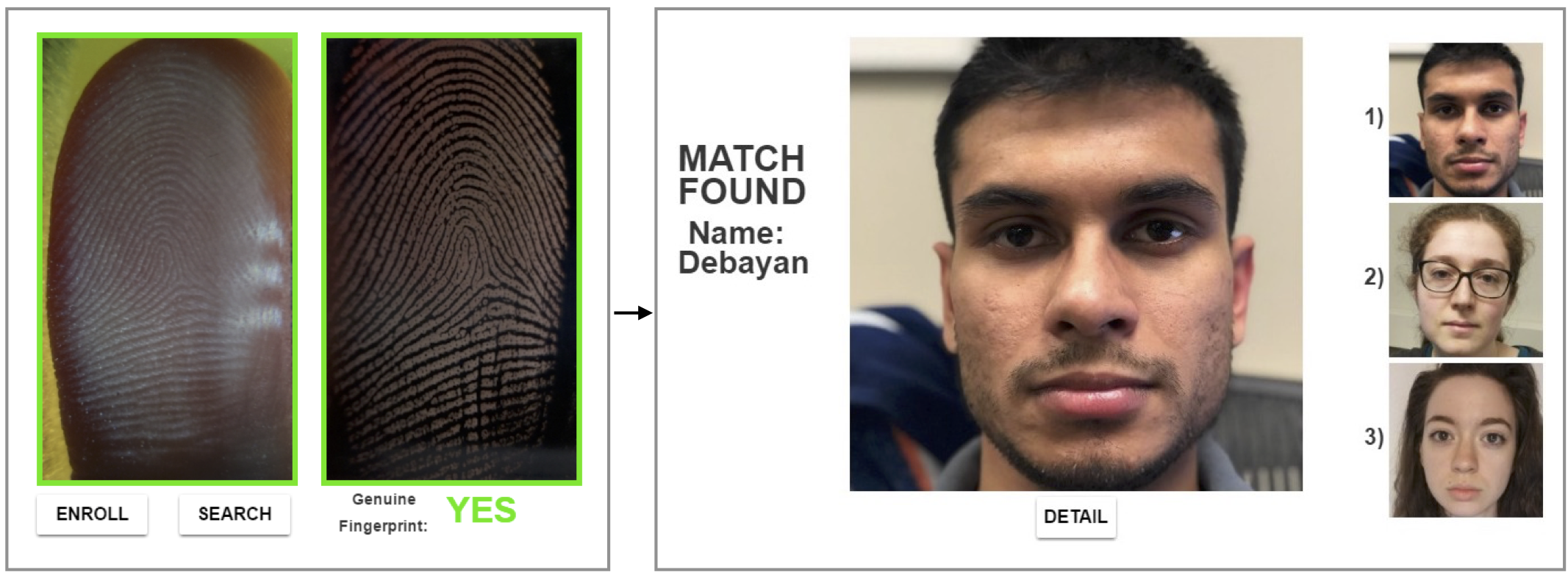}
\caption{The Match in Box GUI is displayed on a 5-inch, 800x480 touch screen.}
\label{fig:gui}
\end{center}
\vspace{-1.5em}
\end{figure*} 

\subsection{Spoof Detector}

The Match in Box spoof detector employs two MobileNet CNNs to classify an input fingerprint image as live or spoof. In particular, one MobileNet CNN is trained to make a binary classification (live or spoof) given an input image from the direct view camera (Fig. \ref{fig:josh_direct}) of Match in Box, and a second MobileNet CNN is trained to make a binary classification (live or spoof) given an input image from the FTIR view camera (Fig. \ref{fig:josh_ftir}) of Match in Box. Finally, the scores generated by each of these respective CNN models are fused using the max rule~\cite{ross2006handbook}.

\subsection{Feature Extractor and Matcher}

Upon completion of fingerprint spoof detection, fingerprint images which are determined to be ``live" are passed to the Match in Box feature extractor. The embedded Match in Box feature extractor operates in two main steps. First, a Convolutional AutoEncoder (CAE) is utilized to locate the $(x, y)$ locations and orientations of the input fingerprint's minutiae points. Then, in a manner similar to~\cite{kai_latent}, a 96x96 patch is cropped around each located minutiae point and passed to a MobileNet CNN to extract a minutiae descriptor. The collection of all extracted minutiae descriptors comprise the template used by the Match in Box embedded matcher. 

To match two templates in Match in Box, pairwise cosine similarity scores are first computed between the sets of minutiae descriptors comprising each template. Using these similarity scores, the top 120 minutiae descriptor pairs are selected and further filtered into a final set $S$ using a graph matching algorithm. Finally, a match score is produced by summing the cosine similarity scores between all pairwise minutiae descriptors in $S$.

\subsection{GUI}

The Match in Box GUI was developed using the popular Electron framework. Electron enables building custom desktop applications using a combination of Javascript, HTML, and CSS. Due to the simplicity, cross-platform deployment capabilities, and modern UI/UX of the framework, it has been extensively used~\cite{electron}. For Match in Box, we specifically develop a light-weight GUI allowing (i) subject enrollment, (ii) verification or search, and (iii) retrieval and display of subject meta-data (Fig. \ref{fig:gui}).

\section{Experimental Analysis}

With the Match in Box fully assembled and operational, we conduct several experiments demonstrating its (i) spoof resistance, (ii) matching performance, and (iii) neonate fingerprint acquisition capability. These experimental results and analyses are further enumerated in the following sections.

\subsection{Spoof Detection}

\begin{figure}[h]
  \centering
  \subfloat[]{\includegraphics[scale=.22]{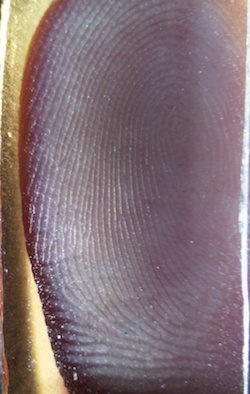}\label{fig:fd}}
  \hfill
  \subfloat[]{\includegraphics[scale=.22]{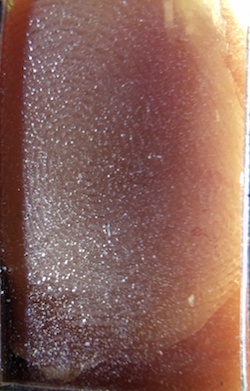}\label{fig:fr}}
  \hfill
    \subfloat[]{\includegraphics[scale=.22]{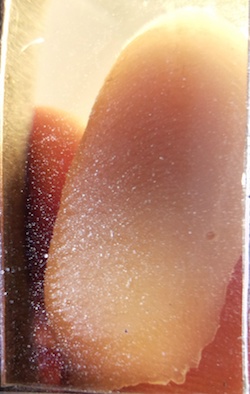}\label{fig:td}}
  \hfill
  \subfloat[]{\includegraphics[scale=.22]{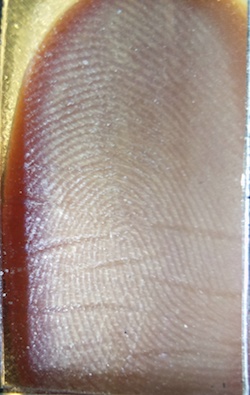}\label{fig:tr}}
  \caption{Match in Box: spoof detection examples. (a) False Detect (live misclassified as spoof) due to the excessively pink color of the subject's finger. (b) False Reject (spoof misclassified as live) due to the close approximation in color of the spoof to human skin. (c, d) True Detect (spoof classified as spoof) and True Reject (live classified as live) examples, respectively.}
  \label{fig:spoof_examples}
 \vspace{-1.0em}
\end{figure}

To demonstrate the spoof detection capability of Match in Box, we first constructed a dataset of both live and spoof finger impressions (Tables 3, 4) (Fig. \ref{fig:spoof_examples}). In particular, we fabricated nearly 1,800 spoofs from three different materials (clear ecoflex, flesh-pigmented ecoflex, and 2D printed paper). The spoof materials were carefully chosen based upon their nominal optical and mechanical similarity to the human skin. In addition, we collected live finger impressions from a diverse population (race, occupation, age) of 56 human subjects. Finally, the live finger impressions and spoof finger impressions were split into 5 training (80\% of impressions) and testing (20\% of impressions) splits. Using these training and testing splits in conjunction with the Match in Box spoof detector, we demonstrate the robust spoof detection capability of Match in Box by computing a mean True Accept Rate (TAR) of \textbf{98.46\%} over the 5-folds at a False Accept Rate (FAR) of \textbf{0.2\%} (standard deviation is \textbf{2.5\%}). Note, the time to perform spoof detection on Match in Box is approximately 120 milliseconds due to the fast inference time available through the on-board VPU.

  \begin{table}[h]
 \centering
\begin{threeparttable}
\begin{tabular}{ |c||c|c|}
 \multicolumn{3}{c}{Table 3: Summary of Spoof Dataset} \\
 \hline
 Material & \specialcell{\# Training \\ Impressions} & \specialcell{\# Testing \\ Impressions}\\
 \hline
 \hline
 Clear Ecoflex & 603 & 151\\
 \hline
 Pigmented Ecoflex & 832 & 208\\
  \hline
 2D Paper & 333 & 84\\
  \hline
  \hline
  Total & 1,768 & 443\\
  \hline
\end{tabular}
\end{threeparttable}
\end{table}

\begin{table}[h]
 \centering
\begin{threeparttable}
\begin{tabular}{ |c||c|c|}
 \multicolumn{3}{c}{Table 4: Summary of Live Dataset} \\
 \hline
 Subjects & \specialcell{\# Training \\ Impressions} & \specialcell{\# Testing \\ Impressions}\\
 \hline
 \hline
 56\tnote{\textdagger} & 2,240 & 560\\
  \hline
\end{tabular}
\begin{tablenotes}
\item[\textdagger] For each subject, 5 impressions were captured from all 10 fingers.
\end{tablenotes}
\end{threeparttable}
\end{table}

\subsection{Matching}

Both authentication (1:1 comparison) and identification (1:N search) performances are reported for Match in Box. As a baseline, we also report the authentication and identification performance when using the Match in Box fingerprint images in conjunction with the well-known and popular COTS Innovatrics SDK~\cite{in}.

Using the fingerprint impressions enumerated in Table 4, we compute the ROC for authentication and the CMC for identification from 5,600 genuine scores and 156,520 imposter scores. From the ROC, we extract True Accept Rates of \textbf{98.0\%} and \textbf{98.57\%} at a FAR of 0.01\%, using Match in Box and Innovatrics, respectively (Table 5). From the CMC, we extract rank one identification rates of \textbf{99.1\%} and \textbf{98.9\%}, using Match in Box and Innovatrics, respectively (Table 6). Both of these results, while reported on a small database, demonstrate the ability of Match in Box to perform state-of-the-art fingerprint authentication and identification. Note, time to perform authentication on Match in Box is approximately 600 milliseconds and time to perform identification from a gallery of 10,000 subjects is less than 2 minutes. 

\begin{table}[h]
 \centering
\begin{threeparttable}
\begin{tabular}{ |c||c|c|}
 \multicolumn{3}{c}{Table 5: Verification Performance} \\
 \hline
 Matcher & \specialcell{False Accept Rate} & \specialcell{True Accept Rate}\\
 \hline
 \hline
 Innovatrics & 0.1\% & 99.0\%\\
  \hline
  Innovatrics & 0.01\% & 98.6\% \\
  \hline
  \hline 
  Match in Box & 0.1\% & 98.6\%\\
  \hline
  Match in Box & 0.01\% & 98.0\% \\
  \hline
\end{tabular}
\end{threeparttable}
\label{tbl:verif}
\end{table}

\begin{table}[h]
 \centering
\begin{threeparttable}
\begin{tabular}{ |c||c|c|}
 \multicolumn{3}{c}{Table 6: Identification Performance} \\
 \hline
 Matcher & \specialcell{Rank-1 \\Identification Rate} & \specialcell{Rank-5 \\Identification Rate}\\
 \hline
 \hline
 Innovatrics & 98.9\% & 99.5\% \\
  \hline
  \hline
  Match in Box & 99.1\% & 99.5\% \\
  \hline
\end{tabular}
\end{threeparttable}
\end{table}

\begin{figure}[t]
  \centering
  \subfloat[]{\includegraphics[scale=.25]{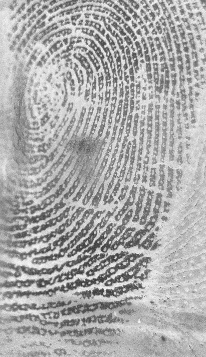}\label{fig:imp1}}
  \hfill
  \subfloat[]{\includegraphics[scale=.25]{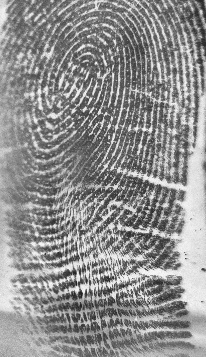}\label{fig:imp2}}
  \hfill
   \subfloat[]{\includegraphics[scale=.25]{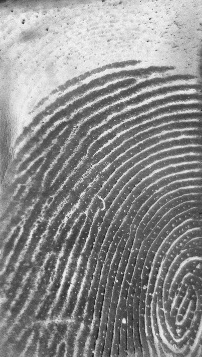}\label{fig:fr1}}
  \hfill
  \subfloat[]{\includegraphics[scale=.25]{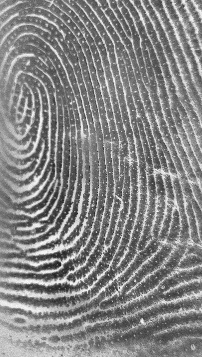}\label{fig:fr2}}
  \caption{Match in Box: failure matching cases. (a, b) Fingerprint image pair resulting in a False Accept, due to close similarity between the imposter fingerprints. (c, d) Fingerprint image pair resulting in a False Reject due to capturing different portions of the fingerprints in consecutive captures. }
  \label{fig:matching_examples1}
 \vspace{-1.0em}
\end{figure}

\begin{figure}[t]
  \centering
  \subfloat[]{\includegraphics[scale=.25]{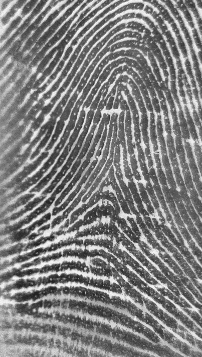}\label{fig:imp1}}
  \hfill
  \subfloat[]{\includegraphics[scale=.25]{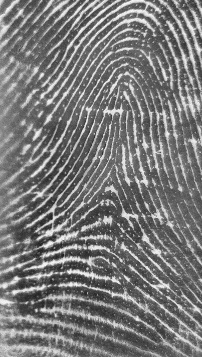}\label{fig:imp2}}
  \hfill
   \subfloat[]{\includegraphics[scale=.25]{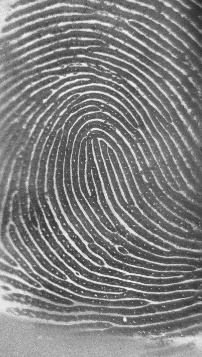}\label{fig:fr1}}
  \hfill
  \subfloat[]{\includegraphics[scale=.25]{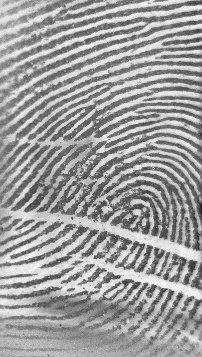}\label{fig:fr2}}
  \caption{Match in Box: successful matching cases. (a, b) Fingerprint image pair resulting in a True Accept. (c, d) Fingerprint image pair resulting in a True Reject. }
  \label{fig:matching_examples2}
 \vspace{-1.0em}
\end{figure}

\subsection{Neonate Fingerprints}

\begin{figure*}[t]
\begin{center}
\includegraphics[clip, trim=0cm 13.5cm 0cm 0cm, scale=0.5]{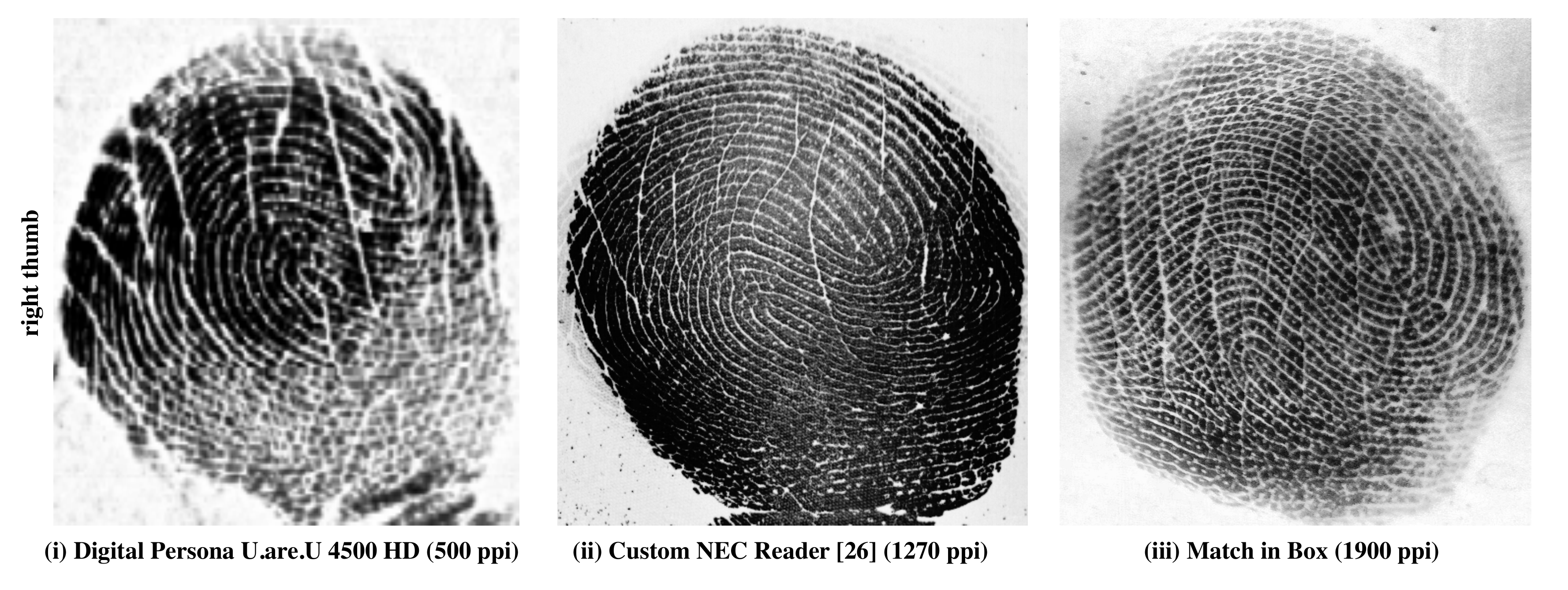}
\caption{Fingerprint images of the right thumb of a 3 month old infant captured with (i) the 500 ppi Digital Persona U.are.U 4500 HD, (ii) a custom, NEC designed, 1270 ppi fingerprint reader for neonate fingerprint acquisition~\cite{infantfingerprint}  and (iii) our 1900 ppi Match in Box.
}
\label{fig:baby}
\end{center}
\vspace{-1.5em}
\end{figure*} 

We demonstrate the utility of Match in Box for collecting high quality neonate fingerprints due to its 1900 ppi cameras. To conduct this experiment, we capture the right thumb print of a 3-month old infant on (i) the 500 ppi Digital Persona U.are.U 4500 HD, (ii) a custom 1270 ppi fingerprint reader designed by NEC~\cite{NEC} for neonate fingerprint acquisition~\cite{infantfingerprint}, and (iii) our proposed 1900 ppi Match in Box. As is shown in (Fig.~\ref{fig:baby}~(i)), the 500 ppi Digital Persona is not able to capture all the minute details of the infant's fingerprints. The 1270 ppi, custom NEC reader performs well in capturing a high quality fingerprint image, however, closer inspection reveals lack of separation between the ridges and valleys around the peripheral of the fingerprint (Fig.~\ref{fig:baby}~(ii)). In contrast, the 1900 ppi Match in Box (Fig.~\ref{fig:baby}~(iii)) is able to capture the entire infant fingerprint, with clean separation between the ridges and valleys throughout the fingerprint image. Furthermore, Match in Box is able to capture more pore information than the NEC reader. 

\begin{figure}[t]
  \centering
  \subfloat[]{\includegraphics[scale=.435]{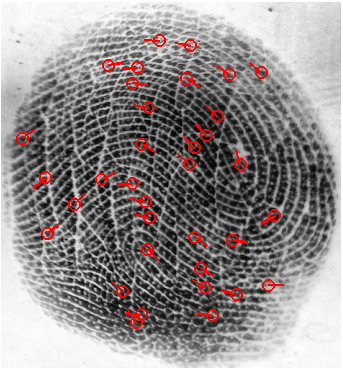}\label{fig:mib1}}
  \hfill
  \subfloat[]{\includegraphics[scale=.45]{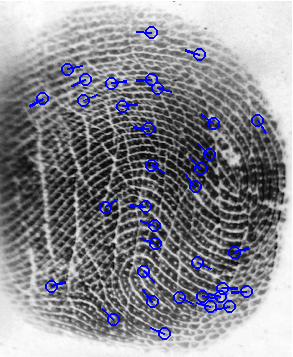}\label{fig:mib2}}
  \caption{Fingerprint images of the right thumb of a 3 month old infant captured by Match in Box (a, b). Verifinger 6.3 was used to generate a match score of 161 between (a) and (b). The score threshold of Verifinger at 0.01\% FAR is 33.   }
  \label{fig:infant_ex1}
 \vspace{-1.0em}
\end{figure}

\begin{figure}[t]
  \centering
  \subfloat[]{\includegraphics[scale=.415]{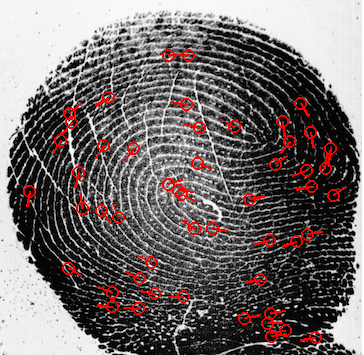}\label{fig:nec1}}
  \hfill
  \subfloat[]{\includegraphics[scale=.4]{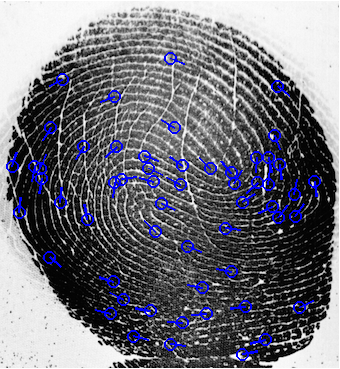}\label{fig:nec2}}
  \caption{(a, b) Fingerprint images of the right thumb of a 3 month old infant captured by NEC~\cite{NEC} (a, b).  Verifinger 6.3 was used to generate a match score of 107 between (a) and (b). The score threshold of Verifinger at 0.01\% FAR is 33.}
  \label{fig:infant_ex2}
 \vspace{-1.0em}
\end{figure}

Finally, we generate match scores between multiple impressions of the infant's fingerprint as captured by Match in Box and the custom NEC reader (Figs. \ref{fig:infant_ex1}, \ref{fig:infant_ex2}). Using Verifinger 6.3, match scores of 161 and 107 are computed for the fingerprint image pairs acquired by Match in Box and the NEC reader, respectively (score threshold of Verifinger at 0.01\% FAR is 33). The annotated minutiae points in Figures \ref{fig:infant_ex1} and \ref{fig:infant_ex2} indicate that the high resolution Match in Box fingerprints are less susceptible to spurious minutiae than the NEC reader fingerprints.

While our initial experiments are qualitative, we posit that our early findings show tremendous promise for utilizing Match in Box for state-of-the-art neonate fingerprint recognition, an area of increasing interest around the world by governments, international agencies, and NGOs alike.

\section{Conclusion}

We have designed and prototyped a complete, spoof resistant, 1900 ppi, fingerprint recognition system, called Match in Box, embedded within a portable form factor. Match in Box is easy and cost-effective to reproduce given the open source software and off-the-shelf components used to construct the device. Our experimental results demonstrate that Match in Box is able to achieve fast, state-of-the-art fingerprint spoof detection, authentication, and identification on device. We have also shown the utility of Match in Box for neonate fingerprint recognition. In the future, we will field test Match in Box at a site in India, where we will specifically target capturing neonate fingerprints of a large number of subjects in a longitudinal study. In doing so, we hope Match in Box will push the boundaries of state-of-the-art neonate fingerprint recognition. We also plan to add a face camera and face matcher to Match in Box, enabling multi-modal biometric fusion and recognition\footnote{This research was supported by the Office of the Director of National Intelligence (ODNI), Intelligence Advanced Research Projects Activity (IARPA), via IARPA R\&D Contract No. 2017 - 17020200004. The views and conclusions contained herein are those of the authors and should not be interpreted as necessarily representing the official policies, either expressed or implied, of ODNI, IARPA, or the U.S. Government. The U.S. Government is authorized to reproduce and distribute reprints for governmental purposes notwithstanding any copyright annotation therein.}.

{\small
\bibliographystyle{ieeetr}
\bibliography{egbib}
}

\end{document}